\documentclass[letterpaper,10pt,conference]{ieeeconf}

\IEEEoverridecommandlockouts
\usepackage{amsmath,amssymb}
\usepackage{booktabs}
\usepackage{graphicx}
\usepackage{multirow}
\usepackage{float}
\usepackage{url}
\usepackage[hidelinks]{hyperref}
\newcommand{\method}{AMC}
\newcommand{\R}{\mathbb{R}}
\newcommand{\sphere}{\mathbb{S}}
\newcommand{\norm}[1]{\left\lVert #1\right\rVert}

\newcommand{\lse}{\operatorname{LSE}}

\title{\LARGE \bf Atomic Motion Coordinate for Language-Steerable and\\
Force-Responsive Manipulation}

\author{Jiaqi Zhai$^{1,11}$, Jingkai Zhao$^{1}$, Chen Yang$^{1}$, Siyuan Ma$^{2}$,\\
Yutian Zhang$^{3,4}$, Liwen Yang$^{3,4}$, Qinglian Wu$^{5}$, Weiqi Fan$^{6}$,\\
Yifei Wang$^{7}$, Yi Zheng$^{8}$, Chenxi Gu$^{9}$, Dong Wei$^{1,\dagger}$,
Wei Zhang$^{10,\dagger}$\\[0.8em]
\normalsize $^{1}$Hangzhou DEEP Robotics Technology Co., Ltd. (DEEP Robotics), Hangzhou, China\\[2pt]
\normalsize $^{2}$Tsinghua University, Beijing, China\\[2pt]
\normalsize $^{3}$Zhejiang University, Hangzhou, China\\[2pt]
\normalsize $^{4}$Shanghai Artificial Intelligence Laboratory, Shanghai, China\\[2pt]
\normalsize $^{5}$Harbin Institute of Technology, Harbin, China\\[2pt]
\normalsize $^{6}$Hainan University, Haikou, China\\[2pt]
\normalsize $^{7}$University of Wisconsin--Madison, Madison, WI, USA\\[2pt]
\normalsize $^{8}$Northwestern University, Evanston, IL, USA\\[2pt]
\normalsize $^{9}$National University of Singapore, Singapore\\[2pt]
\normalsize $^{10}$Eastern Institute of Technology, Ningbo, China\\[2pt]
\normalsize $^{11}$Technical University of Munich, Munich, Germany%
\thanks{$^{\dagger}$Co-corresponding authors.}}

\begin{document}
\raggedbottom
\maketitle
\thispagestyle{empty}
\pagestyle{empty}

\begin{abstract}
\textbf{Can changing only the language instruction redirect a vision-language-action (VLA) policy's end effector, or does the visually driven motion prior dominate?}
We present Atomic Motion Coordinate (\method{}), a geometry-grounded coordinate for steerable and force-responsive manipulation. Each arm
owns thirteen signed translation, rotation, and hold atoms grounded from text
and forward kinematics with vision withheld, and the coordinate is injected into every action-expert
block via a cross-attention residual. Contact history modulates the same
coordinate through a bounded spherical residual that is recomputed from a fixed nominal
latent to regenerate only the unexecuted horizon suffix.
Across 7,520 per-policy offline horizon interventions, opposite-atom separation reaches
92.5/83.1\% (single/dual) versus 39.1/24.0\% for LA4VLA-style. Across 50 real-robot trials per task, AMC raises out-of-distribution (OOD) fruit progress from
60.5\% to 87.8\%; force adaptation raises Plug/Vase from 59.0/71.6\% to
78.5/75.2\%.
\end{abstract}

\section{INTRODUCTION}

Generalist vision--language--action (VLA) policies such as OpenVLA, $\pi_0$,
and $\pi_{0.5}$ turn instructions and camera streams into continuous control
across a widening range of manipulation tasks~\cite{kim2024openvla,black2024pi0,pi2025pi05}.
Demonstrations couple what is said, seen, and done, so high task success can
still reflect a weak language prior over a visually driven routine. Recent work
finds that task targets are learned more readily from visual specifications than
from text, and proposes sparse point flows as a more learnable instruction
channel~\cite{li2024vip}. Rather than replacing language, we ask what it controls.
\textbf{With the same observation and robot state,
does changing only the instruction move the predicted end effector in the
requested direction rather than continue the demonstrated motion?}

We therefore freeze observation, robot state, and inference settings while varying
only the prompt. A fine-tuned $\pi_{0.5}$ does respond---pretraining supplies
common-object semantics and demonstrations supply action support---but
counterfactual trajectories can remain weakly separated or bend through the
native-task direction before diverging. Existing remedies relabel data or steer
internal activations, yet leave the instruction an implicit condition with no
named motor meaning~\cite{glossop2025cast,lin2026la4vla,haon2025mechanistic,wuforethought2025,fang2026visionoverride}.

\begin{figure*}[t]
\expandafter\def\csname @captype\endcsname{figure}
\centering
\makebox[0.98\textwidth]{%
\includegraphics[height=0.43\textwidth]{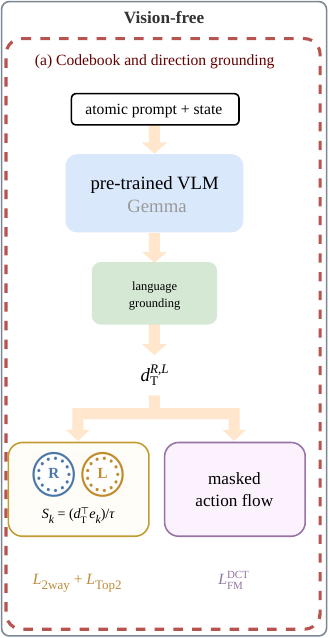}%
\hfill
\includegraphics[height=0.43\textwidth]{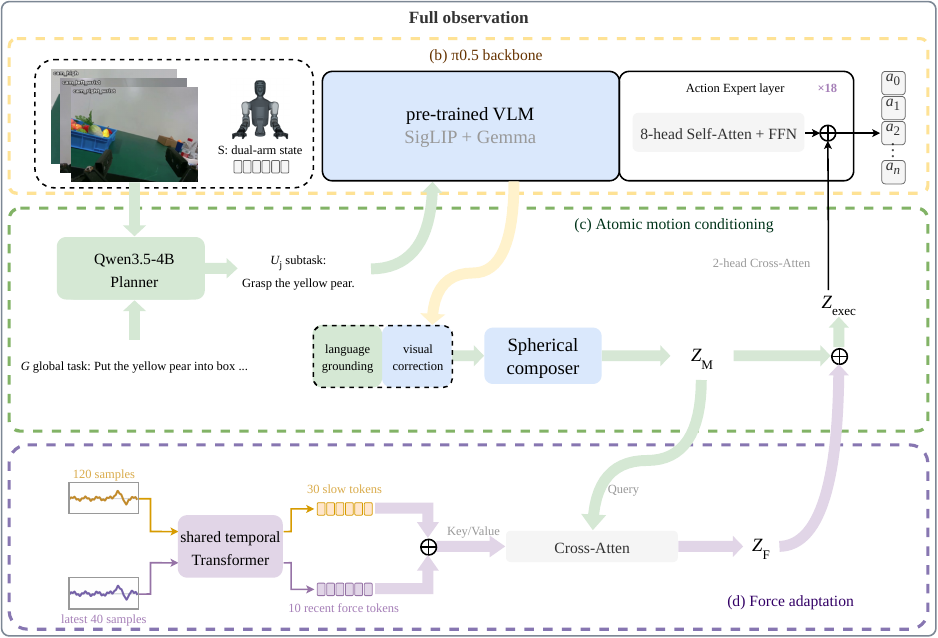}}
\caption{Overview of \method{} and its force-enabled extension. (a) Vision-free codebook and direction grounding.
(b) $\pi_{0.5}$ backbone. (c) Atomic motion conditioning. (d) Force adaptation.
Panels (b,c) support full-observation training and inference; the force-enabled
variant adds (d). Losses in (b)--(d) are omitted for clarity.}
\label{fig:overview}
\end{figure*}

\method{} (Fig.~\ref{fig:overview}) grounds per-arm motion coordinates from
atomic language and forward-kinematics (FK) supervision without vision. Each
arm owns a spherical codebook of thirteen atoms: signed translations, signed
rotations, and hold. Atomic text and normalized joint state condition the
masked 50-step action flow and its low-frequency structure, tying atom identity
to measured motion rather than object appearance. Under full observation,
the coordinate specifies a motion trend, while the action expert generates
its scene-conditioned realization. A bounded tangent component adds contextual
detail, and the ordered right/left pair conditions all 18 action-expert blocks
through an independent, zero-initialized cross-attention path. The original
self-attention and flow-time paths remain structurally intact, and the route is
initially identity-preserving. The planner selects a subtask from a predefined
vocabulary using a fixed prompt template; it outputs neither atoms nor weights.
The low-level policy maps that subtask, images, and state to the motion coordinate.

Contact feedback adapts the ongoing motion without requiring a new semantic
instruction. A shared encoder converts one second of wrench and state history
into 30 slow tokens and the newest post-anchor samples into 10 fast tokens.
The nominal $z_M$, augmented with the current real-time chunking (RTC)
phase~\cite{black2025trainingrtc}, queries this memory to produce the force
correction $z_F$. A bounded spherical update regenerates only the unexecuted
suffix of the 50-step horizon. Recomputing each update from the fixed $z_M$
avoids accumulating corrections across contact updates.

Two boundaries scope our claim: we do not argue that a strong baseline ignores
language, nor that an explicit coordinate can ground concepts absent from both
pretraining and robot data. Throughout this paper, OOD denotes
held-out recombinations of known targets, orders, and scene configurations.
Within that scope, AMC raises opposite-atom separation from
39.1/24.0\% to 92.5/83.1\% for single/dual prompts over 7,520 per-policy
matched offline interventions from recorded real-robot states, while retaining
closed-loop competence that steering-oriented baselines substantially lose.

Our contributions are:
\begin{itemize}
    \item \textbf{A geometry-supervised, intervenable language--motion interface.}
    FK-grounded per-arm coordinates connect atomic language to continuous
    action generation and expose a site for inspecting and intervening on
    motion under full observation.
    \item \textbf{A bounded update anchored in the same representation.}
    Force feedback revises the unexecuted action suffix through a bounded,
    non-accumulating spherical correction to the nominal motion coordinate.
    \item \textbf{Validation across three separated capability regimes.}
    Atomic interventions, fruit trials in familiar and held-out layouts, and
    contact-rich trials test geometric-direction steering, closed-loop subtask
    steering, and force adaptation.
\end{itemize}

\section{RELATED WORK}

\subsection{Language-Conditioned VLA and Latent Actions}
OpenVLA predicts discretized actions from a vision-language backbone, whereas
$\pi_0$ and $\pi_{0.5}$ attach a continuous flow action expert to a pretrained
vision-language model (VLM)~\cite{kim2024openvla,black2024pi0,pi2025pi05}, continuing the broader use of chunked,
receding-horizon prediction in ACT and Diffusion
Policy~\cite{zhao2023act,chi2023diffusionpolicy}. Action abstractions have been
learned by residual vector quantization, frequency-domain tokenization, and
visual future prediction~\cite{lee2024vqbet,pertsch2025fast,ye2025lapa}, and
LARY argues that semantic decodability and physical reconstruction must be
scored separately~\cite{nie2026lary}. Our codebook differs in origin: it begins
from an embodiment-specific geometric vocabulary rather than a learned token
inventory, while the deployed policy still emits continuous trajectories.

Closest to us, LA4VLA removes vision during language--action pretraining, then
returns to conventional full-observation VLA training~\cite{lin2026la4vla}.
It changes the curriculum but retains no fixed FK-derived dictionary, per-arm
intervention variable, or contact-update site. We instead ground signed
geometric axes, preserve their coordinates under full observation, and expose
them directly to the action expert. Residual Semantic Steering (RSS)~\cite{stable2026slg} combines Monte Carlo Syntactic Integration with
Residual Affordance Steering to amplify language over a vision-only prior,
but operates at the action-distribution level without a named per-arm coordinate.
CAST and CofactVLA use counterfactual
relabeling~\cite{glossop2025cast,zhang2026cofactvla}; other work steers internal
activations or filters plans through latent-aligned VLM
verification~\cite{haon2025mechanistic,wuforethought2025}. None of these methods
exposes a fixed, per-arm FK-grounded coordinate shared by language intervention
and force adaptation.

\subsection{Hierarchical Planning and Fast Contact Feedback}
Hierarchical policies ground language into executable skills and revise plans
across stages~\cite{ichter2023saycan,feng2025reflective}. We keep
completed-state information in a compact recursive summary from
Qwen3.5-4B~\cite{qwen2026qwen35}, a boundary-rate planner rather than a
controller: only its subtask reaches the motor policy.

On the contact side, RDP, ForceVLA, and FAVLA pair a slow visuomotor route with
faster tactile or force conditioning~\cite{xue2025rdp,yu2025forcevla,li2026favla}.
AT-VLA and T-Rex route native-rate signals through specialized action
paths~\cite{li2026atvla,niu2026trex}, while predictive models forecast tactile
evolution before revising actions~\cite{zhou2026touchworld,zhang2026retouch}.
FACTR shows that visually dominant policies can underutilize force without an
explicit force-attending curriculum~\cite{liu2025factr}.
We instead place force correction in the same spherical space as language,
recomputed from one fixed nominal latent.

\section{METHOD}

\method{} comprises three functional components: a planner that converts a
global task into one executable subtask at a time (Sec.~\ref{sec:hierarchy});
a per-arm atomic coordinate that is first grounded from language and
proprioception with vision withheld, then reused under full observation to
condition every depth of the action expert
(Sec.~\ref{sec:vocab}--\ref{sec:zm}); and a spherical force update that
revises the coordinate during contact without displacing its semantic anchor
(Sec.~\ref{sec:force}). The low-level coordinate pathway is trained
progressively through vision-free grounding, full-observation continuation,
and force-adaptation stages.

\subsection{Hierarchy and Action Interface}
\label{sec:hierarchy}
At each 1\,Hz tick, the upper loop receives global task $G$, the current
observation $O_j$, previous memory $M_{j-1}$, and previous subtask
$U_{j-1}$. Qwen3.5-4B emits
\begin{equation}
(M_j,U_j)\sim p_\psi(\cdot\mid G,O_j,M_{j-1},U_{j-1}).
\label{eq:planner}
\end{equation}
$M_j$ is a compact cross-frame temporal anchor recording completed events and
persistent object state. $U_j$ is either the previous subtask carried forward
or a new executable command when current visual evidence warrants a
transition. Qwen3.5-4B is used without task-specific supervised fine-tuning
(SFT); a fixed prompt template constrains its output to the predefined
subtask vocabulary. Only $U_j$ is passed to the low-level policy.

The low-level policy observes three RGB images $I_t$, the active subtask
$U_j$, and dual-arm state
\begin{equation}
s_t=[q_t^L,g_t^L,q_t^R,g_t^R]\in\R^{16}.
\end{equation}
It predicts $H=50$ actions at 30 Hz through the 32-D $\pi_{0.5}$ interface;
the first 16 coordinates control the two seven-DoF arms and grippers, while
the remaining coordinates preserve base-checkpoint compatibility. The action
expert retains the standard $\pi_{0.5}$ conditional flow-matching
target~\cite{pi2025pi05}.

One subtask typically spans several 50-step horizons; during training each
arm--horizon receives either a strict single/dual atomic target or, when no
strict atom is supported, a weighted Top-5 FK target used only during
full-observation training. A subtask is thereby realized as a sequence of
per-horizon motor targets (Sec.~\ref{sec:vocab}). At inference, $U_j$, images,
and state drive the low-level Q1/Q3 heads to predict codebook-aligned per-arm
directions; the resulting inspectable motor coordinate can subsequently be
revised by force (Sec.~\ref{sec:force}).

\subsection{An Atomic Vocabulary Defined by Forward Kinematics}
\label{sec:vocab}
The labeling rule must be immune to the language--motion correlation that we
seek to break. We therefore derive every atomic target from measured geometry,
never from subtask text. For each arm $a\in\{L,R\}$,
\begin{equation}
\mathcal A=\{\pm x,\pm y,\pm z,\pm r_x,\pm r_y,\pm r_z,\mathrm{hold}\}.
\end{equation}
The first six atoms denote signed translations and the next six signed
rotations in the robot base frame. We compute end-effector motion with the deployment URDF and a 0.20-m
tool-center-point (TCP) offset, using signed translation and the spatial
rotation log of $R_hR_0^\top$ over each horizon. A horizon dominated by one
component receives a single positive atom. When two compatible components
jointly dominate, such as in a diagonal reach, it receives a weighted Top-2
target; forcing it onto one axis would assign a direction the arm did not take.
Top-2 weighted labels are used in vision-free grounding. During
full-observation training, an arm without a strict single- or dual-atom label
instead uses a normalized Top-5 FK mixture target (Sec.~\ref{sec:zm}), without
assigning an unsupported discrete atom. Because labels are derived from FK
alone, object-aware phrasing cannot alter the supervision targets.

Each arm owns a normalized 512-D dictionary
\begin{equation}
E^a=\{e_1^a,\ldots,e_{13}^a\},\qquad e_k^a\in\sphere^{511}.
\end{equation}
Here, $d_T^a$ denotes the vision-free, language-and-proprioception-grounded
unit direction for arm $a\in\{R,L\}$. For positives $P$ and all other atoms $N$,
$\rho_k=(d_T^a)^\top e_k^a$, and $s_k=\rho_k/\tau_a$ is its
temperature-scaled cosine similarity to atom $k$, with $\tau_a=0.10$. The sample-wise
multi-positive Two-Way loss is
\begin{equation}
\mathcal L_{\mathrm{2way}}^a=
\operatorname{softplus}\!\left[
\lse_{p\in P}(-s_p)+\lse_{n\in N}(s_n)\right].
\label{eq:twoway}
\end{equation}
The two log-sum-exp terms push the weakest positive up and the strongest
negative down within the same sample, so a dual-motion horizon can keep both of
its atoms; a softmax objective over a single label would instead force the two
to compete. For FK mixture weights $w_p\geq0$ with $\sum_{p\in P}w_p=1$, we
define
\begin{equation}
\begin{aligned}
u_P^a &= \frac{\sum_{p\in P}w_pe_p^a}{\norm{\sum_{p\in P}w_pe_p^a}},\\
\mathcal L_{\mathrm{Top2}}^a &= 1-(d_T^a)^\top u_P^a.
\end{aligned}
\label{eq:top2-cosine}
\end{equation}
This weighted-cosine term aligns a single or dual direction with its normalized
FK code mixture.

\subsection{Vision-Free Atomic Grounding}
\label{sec:zt}
This stage jointly grounds the atomic codebooks and motion directions using
the $\pi_{0.5}$ VLA with atomic modules but no visual input
(Fig.~\ref{fig:overview}(a)). Withholding vision is the point of this phase,
not a simplification. A policy
that can see the scene may satisfy an atomic prompt by replaying whatever
motion that scene usually elicits, leaving the atom no independent meaning. We
therefore supply only the atomic prompt and the normalized robot state,
encoded as the pretrained VLM state prefix. This removes the direct scene
shortcut and forces the atomic instruction to disambiguate the desired motion
direction given the current state. The Q1 and Q3 heads emit unit directions
$d_T^R,d_T^L$ that condition the same action expert used at deployment. We mask
the horizon flow loss to the labeled arm's joints, leaving grippers, the
opposite unlabeled arm, and padding coordinates untargeted, so a direction is
never credited for motion it did not name.

To suppress 30-Hz action noise while preserving coarse motion, the flow
matching target for the labeled arm is replaced by a discrete cosine transform (DCT-II) low-frequency
reconstruction: $C^a=D_{0:K-1}Y^a$ with $K=3$ (DC term and two lowest
non-DC modes~\cite{pertsch2025fast}) extracts three low-frequency coefficients;
the length-50 reconstruction $\tilde{Y}^a=D_{0:K-1}^\top C^a$ then replaces the raw flow target.
Gripper and unlabeled channels remain
masked. This gives $\mathcal L_{\mathrm{FM}}^{\mathrm{DCT}}$, and the
complete grounding objective is
\begin{equation}
\mathcal L_T=\mathcal L_{\mathrm{FM}}^{\mathrm{DCT}}
+0.10\sum_a\mathcal L_{\mathrm{2way}}^a
+0.10\sum_a\mathcal L_{\mathrm{Top2}}^a.
\label{eq:zt}
\end{equation}
Thus atom identity is tied to both a 50-step physical action and its compact
low-frequency structure, without image features explaining the label. At the
end of this 25k-update grounding stage, the codebooks are frozen for
full-observation continuation.

\subsection{Full-Observation Motion Latent}
\label{sec:zm}
Under full observation (Fig.~\ref{fig:overview}(b,c)), the VLM receives the
three camera views, robot state, and active subtask $U_j$. The atomic coordinate
sets a motion trend, and the action expert supplies its scene-conditioned
realization. Bounded detail accommodates scene-dependent variation not specified
by the atomic direction. Q1/Q3 predict right/left arm directions and Q2/Q4 their
tangent-plane details. For arm $a$, the direction query yields
$d_M^a\in\sphere^{511}$, while the detail query yields four bounded coefficients
$c^a=\tanh(W_ch_{\mathrm{detail}}^a)$. We project four learned seeds into the
tangent plane of $d_M^a$ and orthonormalize them into the four-vector basis
$B_\perp^a=\{B_{\perp,k}^a\}_{k=1}^{4}$. Let
$r_M^a=0.5\sum_{k=1}^{4}c_k^aB_{\perp,k}^a$. Writing
$r_\perp=(I-bb^\top)r$ for the tangent projection, the bounded spherical update is
\begin{align}
\theta&=\theta_{\max}\tanh\!\left(\frac{\lVert r_\perp\rVert}{\theta_{\max}}\right)\notag\\[-1pt]
\mathcal S(b,r;\theta_{\max})
&=\operatorname{Norm}\!\left(
\cos\theta\,b+\sin\theta\,\frac{r_\perp}{\lVert r_\perp\rVert}\right).
\label{eq:spherical-update}
\end{align}
Here $b\in\sphere^{511}$, $\operatorname{Norm}$ is $\ell_2$ normalization, and
the zero-residual case is defined by continuity as $\mathcal S(b,0;\theta_{\max})=b$
(the implementation uses a numerical stabilizer). Although $r_M^a$ is tangent
by construction, the generic projection is retained because the same update is
reused for the force residual in Sec.~\ref{sec:force}.
The full-observation coordinate is
$z_M^a=\mathcal S(d_M^a,r_M^a;\pi/4)\in\sphere^{511}$.
Strict single/dual horizons use stopped-gradient $d_T^a$ as teacher. A
non-strict horizon instead uses the frozen FK mixture
$u_{T_5}^a=\operatorname{Norm}(\sum_{p\in T_5^a}w_pe_p^a)$ from its five largest
normalized components, with loss $1-(d_M^a)^\top u_{T_5}^a$; no unsupported
atom is assigned. A 0.005-weighted $20^\circ$ free cone penalizes geodesic
excess between $d_M^a$ and $z_M^a$. The $45^\circ$ cap bounds detail around
$d_M^a$, while teacher alignment anchors it to the grounded target. Thus
full-observation contextual detail can shape the trajectory without changing
latent norm. The ordered pair $[z_M^R,z_M^L]$ conditions every
action-expert block through a separate two-head cross-attention residual:
\begin{equation}
H_\ell\leftarrow H_\ell+
\operatorname{CA}_\ell\!\left(H_\ell,[z_M^R,z_M^L]\right),
\quad \ell=1,\ldots,18.
\label{eq:latent-conditioning}
\end{equation}
Its output projection is zero-initialized. The backbone's eight-head
self-attention, feed-forward blocks, and flow-time AdaRMS remain unchanged.

During final stabilization, the direction supplied to the action expert is
replaced by its stopped-gradient teacher ($d_T$ for atomic labels and the
frozen Top-5 mixture otherwise), while $d_M$ remains cosine-supervised to that
teacher. This prevents early visual drift from destabilizing action learning;
the tangent detail remains predicted from full observation. At inference, the
learned $d_M$ is used without teacher substitution.

The flow objective alone does not guarantee strong utilization of the coordinate:
the pretrained action expert can initially reduce the loss while largely
bypassing the zero-initialized cross-attention path.
We therefore add a matched counterfactual pass: a second run replaces one
supervised arm direction with a hard-negative codebook atom outside that arm's positive support,
holding every other input and the flow time fixed. With per-arm flow losses $\ell_{\mathrm{corr}}$ and
$\ell_{\mathrm{wrong}}$,
\begin{equation}
\mathcal L_{\mathrm{cf}}=
\left[m+\ell_{\mathrm{corr}}-\ell_{\mathrm{wrong}}\right]_+,
\quad m=2\times10^{-3}.
\label{eq:counterfactual}
\end{equation}
\vspace*{1pt}
The wrong direction must therefore reconstruct the demonstration strictly worse
than the correct one by a margin $m$; once that gap holds, the hinge is
inactive and the term stops competing with the flow objective. The final
continuation weights it at 0.10, alongside the full flow loss, cosine alignment
between the predicted direction and its stopped-gradient teacher, and
Top-5 weighted-cosine supervision.

\subsection{Force-Responsive Latent Modulation}
\label{sec:force}
The force-enabled variant augments this full-observation policy with the
force-adaptation branch (Fig.~\ref{fig:overview}(d)). The force branch predicts
a residual $z_F$ that adapts nominal $z_M$ to contact without requiring a new
subtask instruction. For a 50-step rollout, RTC offset
$i\in\{0,10,20,30,40\}$ denotes the number of already committed actions at a
decision boundary.
Each wrist provides a calibrated six-axis wrench at 120 Hz. Per arm,
non-overlapping groups of four consecutive force--state samples are projected
into 512-D tokens (120\,Hz$\to$30\,Hz), then passed through a shared two-layer
temporal transformer and an arm-specific projection. The newest 120 raw samples
yield 30 slow tokens $F_{\mathrm{slow}}^a\in\R^{30\times512}$. At offset $i>0$,
the most recent at most 40 force--state samples acquired after the preceding
decision anchor form 10 fast tokens $F_{\mathrm{fast}}^{a,(i)}\in\R^{10\times512}$;
unavailable fast positions at $i=0$ are validity-masked. The paths share encoder weights and concatenate only in
cross-attention. Before learning the action correction, B1 trains only this
encoder with an auxiliary single-layer GRU decoder: the temporal transformer
compresses slow force--state history into per-arm tokens, and the GRU decodes
them into future force increments, shaping the encoder so slow tokens carry
contact-predictive structure. The GRU is discarded at deployment. B2 freezes
the VLM, nominal motion queries, codebooks, atomic cross-attention, and B1
force encoder; the future predictor is excluded from deployment. It then trains
only the action expert and force adapter under RTC. $z_M^a$ plus a sinusoidal RTC-phase embedding
$\phi_i$ forms a query over the 30+10 token memory. A two-head cross-attention
adapter and zero-initialized output produce one direct 512-D residual:
\begin{align}
q^{a,(i)}&=\operatorname{LN}(W_qz_M^a+\phi_i),\\
z_F^{a,(i)}&=W_o\operatorname{swish}\!\left(
W_h\operatorname{CA}_2(q^{a,(i)},[F_{\mathrm{slow}}^a;
F_{\mathrm{fast}}^{a,(i)}])\right),\\
z_{\mathrm{exec}}^{a,(i)}&=\mathcal S(z_M^a,z_F^{a,(i)};\pi/4).
\label{eq:force}
\end{align}
The update uses the same $20^\circ$ free cone, $45^\circ$ cap, and 0.005 hinge
weight as the visual route. Each $z_F^{a,(i)}$ is recomputed from nominal
$z_M^a$, not added to a previous $z_{\mathrm{exec}}$: this prevents recursive
drift from the named atom and bounds total departure to one residual. The ordered $[z_{\mathrm{exec}}^R,z_{\mathrm{exec}}^L]$
pair replaces the nominal pair in Eq.~\eqref{eq:latent-conditioning} at every
action block.

At training offset $i$, the demonstrated prefix of length $i$ is clamped and
excluded from loss; only the regenerated suffix $i{:}H$ is supervised with
uniform per-step weighting. At deployment, each boundary retains the committed
prefix and regenerates only the unexecuted suffix~\cite{black2025trainingrtc};
force is sampled at 120\,Hz, but policy updates occur once per ten executed
actions ($\approx3$\,Hz).

\subsection{Training and Model Size}
\label{sec:training}
Both recipes begin with 25k vision-free $z_T$ updates, after which codebooks
are frozen (Table~\ref{tab:training}). The force-free recipe uses 25k $z_M$
updates; the force-enabled recipe uses 20k $z_M$ updates followed by B1 and B2.
B1 trains only the force encoder and future-wrench predictor for 5k updates.
B2 trains the action expert and force adapter for 5k RTC updates with the
encoder frozen. Both recipes thus have 25k post-grounding action-expert
updates; B1 adds encoder-only pretraining, not action-expert updates.

\begin{table}[t]
\centering
\vspace*{6pt}
\caption{Training schedule. Steps denote per-stage updates.}
\label{tab:training}
\setlength{\tabcolsep}{3pt}
\footnotesize
\begin{tabular}{llr}
\toprule
Stage & Trainable & Steps \\
\midrule
$z_T$ grounding & VLA + atomic modules & 25k \\
$z_M$ full observation & VLA + atomic modules & 25k / 20k$^{*}$ \\
B1 prediction (force only) & force encoder + predictor & 5k \\
B2 adaptation (force only) & action expert + force adapter & 5k \\
\bottomrule
\end{tabular}
\vspace{1mm}

{\footnotesize
VLA: backbone + action expert. Atomic modules: queries, heads, codebooks, and
$z_M$ cross-attention. Codebooks freeze after $z_T$.
$^{*}$20k $z_M$ for the force-enabled recipe.}
\vspace*{-6pt}
\end{table}

All stages use 8$\times$H800 GPUs, global batch size 256, and cosine annealing.
The $z_T/z_M$ stages use peak lr $2.5\!\times\!10^{-5}$ (minimum
$2.5\!\times\!10^{-6}$; 1000 warmup steps); B1 uses peak lr $10^{-4}$
(minimum $10^{-5}$; 300 warmup steps). B2 uses
$5\!\times\!10^{-6}$ to $10^{-6}$ with 200 warmup steps. The gap in
Table~\ref{tab:main} is not a causal force estimate: Table~\ref{tab:force-ablation}
isolates the route by varying only force content within one checkpoint.

The deployed atomic and force routes add 34.56M and 12.40M parameters,
totaling 46.96M (1.40\% over the 3.353B $\pi_{0.5}$ backbone); the DCT-based target construction
and B1 future-wrench predictor are training-only.

\begin{figure*}[t]
\expandafter\def\csname @captype\endcsname{figure}
\centering

\includegraphics[width=0.98\textwidth]{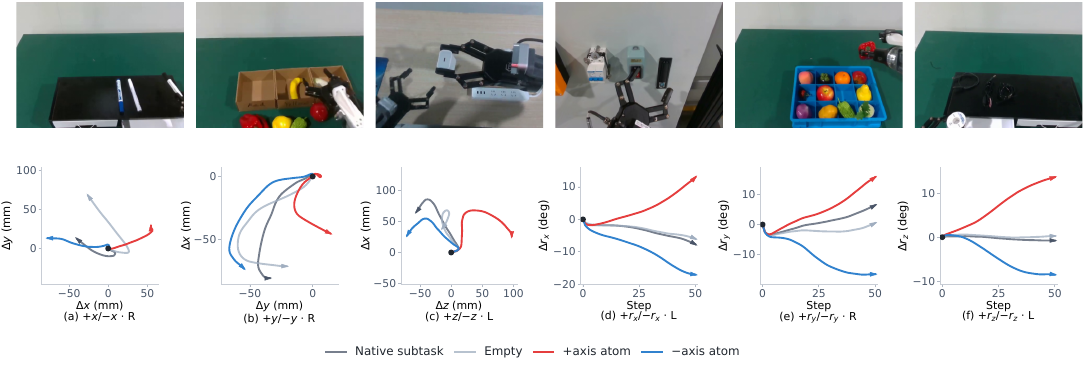}
\caption{Single-atom prompt interventions with fixed images and state. Native,
Empty, and opposite-signed prompts vary one atom. Panels (a)--(c) show
horizon-50 base-frame TCP paths; (d)--(f) show signed rotation trajectories.
Dots and arrowheads mark common starts and endpoints. Panels (a)--(f) use
marker insertion, apple grasp, power-strip hold, switch press, chili placement,
and solder-coil placement states, respectively.}
\label{fig:atomic-steering-qualitative}

\end{figure*}
\section{EXPERIMENTS}

\subsection{Platform, Data, and Four Task Families}
Experiments use a CR1 dual-arm platform with two seven-DoF arms, two grippers,
a base camera, two wrist cameras, and a six-axis force/torque sensor at each wrist. RGB,
state, and actions are recorded at 30 Hz; wrench is recorded at 120 Hz. FK
annotation, evaluation, and visualization use the same URDF and 0.20-m TCP.
The unified training corpus contains 2,375 episodes and 3.07M valid 30-Hz
training samples across cabinet, drawer, fruit, mixed manipulation, screwdriver, plug, and
vase domains. Normalization statistics use the training split only.
On a single RTX~4090, atomic-only inference takes 117.4\,ms; force-enabled
slow/fast loops take 118.7/45.5\,ms, within the 333\,ms ten-action window.

We evaluate four families. Multi-fruit
sorting evaluates closed-loop subtask steering: at annotated semantic
boundaries, an operator selects an object- or destination-specific instruction
from the same predefined execution-subtask vocabulary, and the policy is
evaluated on whether it follows that instruction. Fruit positions are
randomized across familiar configurations.
Drawer retrieval and plug insertion (Plug) requires
opening the drawer, grasping and reorienting
the power adapter, aligning the connector, and completing a contact-rich
insertion, with execution subtasks supplied by the Qwen3.5-4B planner.
OOD fruit ordering tests the same subtask steering under unseen fruit
placement and target-box positions.
Bimanual vase wiping alternates stabilization, rotation, and sustained
wiping contact, with subtasks supplied by the Qwen3.5-4B planner.

\subsection{Baselines and Metrics}
We compare six policies: fine-tuned $\pi_{0.5}$~\cite{pi2025pi05}, matched reproductions of LA4VLA~\cite{lin2026la4vla} and ForceVLA~\cite{yu2025forcevla}, an adaptation of RSS~\cite{stable2026slg} to our real-robot setting, our force-free \method{}, and \method{} + Force.
All reproductions use the same three-view RGB observations, 16-D dual-arm state, 32-D action interface, and evaluation protocol; training subsets are matched to each method's evaluated task families.
For real-robot task progress, LA4VLA and RSS are evaluated only on fruit tasks; ForceVLA only on Plug and Vase.

The primary robot metric is mean normalized task progress. Each method is
evaluated in 50 trials per task. If trial $i$ earns $s_i$ out of the task's
prespecified maximum $S$, its progress is $100s_i/S$. Plug progress uses four
one-point macro-stages: opening the drawer, establishing the grasp on the
adapter, aligning the connector, and completing insertion. The planner may
issue multiple finer-grained execution subtasks within each scored stage. Vase uses five points: two for rotation (one per half-turn, stabilization subsumed),
two for wiping (one per half-clean), and one for returning the vase. Scoring and termination conditions are fixed
across methods. For fruit tasks, an operator selects the execution subtask from
the same predefined vocabulary at annotated semantic boundaries; task progress
measures whether the policy follows it. For Plug and Vase, Qwen3.5-4B generates
subtask prompts automatically. Fruit and contact-task results measure distinct
capabilities and are not pooled into a single average. Atomic
steerability follows the matched protocol of Sec.~\ref{sec:atomic-steering}.
Planner reliability is evaluated separately at subtask boundaries.

\begin{figure*}[t]
\expandafter\def\csname @captype\endcsname{figure}
\centering

\includegraphics[width=0.755\textwidth,height=0.231\textwidth]{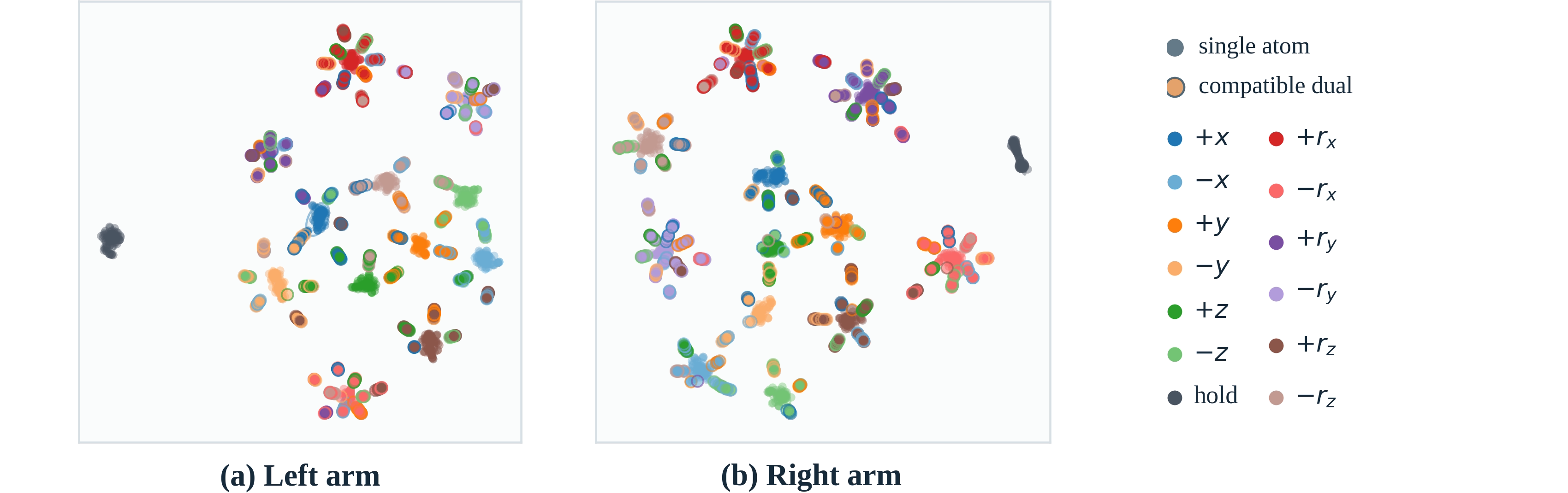}
\caption{Per-arm t-SNE of grounded directions $d_M^a$. Colors mark target atoms;
circles and two-tone circles denote single and dual samples.}
\label{fig:codebook-tsne}
\end{figure*}
\subsection{Atomic Steering}
\label{sec:atomic-steering}

Figure~\ref{fig:atomic-steering-qualitative} makes the counterfactual visible:
with images and state fixed, native trajectories expose the scene-associated
task prior, Empty removes text, and changing only the signed atom separates
endpoints or rotations. Table~\ref{tab:steering} quantifies this contrast. We
cluster standardized 14-D joint states (grippers excluded) into
250 clusters. Retaining outcome-agnostic modes observed at least five times
yields 1,880 cluster--mode pairs (830 single, 1,050 dual) across 948 states
(452 left, 496 right). Four states per pair yield 7,520 interventions per
policy (3,320 single, 4,200 dual). All policies use the same interventions;
within a policy, images, state, normalization, sampler, and initial flow noise
are fixed as prompts change. Empty uses the same images and state with no text;
FK maps predictions to TCP without execution.

The single-atom sweep contains 1,100 left-arm and 2,220 right-arm interventions
per policy. Anchor selection is balanced, but unequal source support means the
pooled rate is not an equal-weight arm average.

A translation or rotation intervention succeeds when its horizon-50 endpoint
moves in the requested signed base-frame direction beyond threshold; a dual
combines two compatible components on one target arm (the other holds) and
requires both. Pair separation instead requires correctly ordered,
threshold-separated signed-atom endpoints, without either endpoint individually
clearing threshold (Table~\ref{tab:steering}). Vs.\ Empty applies the signed
threshold to the prompted endpoint relative to Empty. Pair separation uses the
reversal-supported subset: 2,316 single and 2,224 dual interventions.
Absolute tests requested motion, Vs.\ Empty isolates language beyond the same
visual prior, and pair separation tests opposite-prompt ordering.

The 10 mm/$2^\circ$ threshold acts only as a noise gate: retained interventions
have mean controlled-axis magnitudes of 55.5 mm and $20.2^\circ$.

\begin{table}[t]
\centering

\caption{Matched horizon-50 atomic steering at 10 mm/$2^\circ$ (\%).
AMC uses the preceding non-spherical 50k checkpoint (25k $z_T$ + 25k $z_M$).
The source-support filter is applied to each policy.}
\label{tab:steering}
\setlength{\tabcolsep}{2.4pt}
\footnotesize
\begin{tabular}{llccc}
\toprule
Method & Atom type & Absolute & Vs. Empty & Pair sep. \\
\midrule
\multirow{2}{*}{Fine-tuned $\pi_{0.5}$}
& Single & 50.5 & 18.3 & 18.1 \\
& Dual   & 37.1 & 6.9  & 8.1  \\
\addlinespace[2pt]
\multirow{2}{*}{LA4VLA-style}
& Single & 52.9 & 31.7 & 39.1 \\
& Dual   & 37.9 & 13.2 & 24.0 \\
\addlinespace[2pt]
\multirow{2}{*}{RSS}
& Single & 52.6 & 39.2 & 60.7 \\
& Dual   & 26.2 & 9.7  & 22.5 \\
\addlinespace[2pt]
\multirow{2}{*}{\method{}}
& Single & \textbf{80.2} & \textbf{74.5} & \textbf{92.5} \\
& Dual   & \textbf{57.9} & \textbf{39.1} & \textbf{83.1} \\
\bottomrule
\end{tabular}
\end{table}

Fine-tuned $\pi_{0.5}$ reaches 50.5/37.1\% Absolute success but only
18.3/6.9\% Vs.\ Empty and 18.1/8.1\% pair separation (single/dual): native
motion direction alone does not establish language control. The gap shows that
a correct absolute motion can still follow the state- and scene-conditioned
prior rather than the prompted atom. LA4VLA-style adds
21.0/15.9 points of pair separation over $\pi_{0.5}$, but only 2.4/0.8 points
of Absolute success and does not preserve closed-loop fruit progress
(Table~\ref{tab:main}). RSS has the strongest baseline single-atom pair
separation (60.7\%), yet its dual Absolute/Vs.\ Empty/pair scores fall to
26.2/9.7/22.5\%; this does not isolate residual interference as the cause.
\method{} leads all metrics: 80.2/57.9\% Absolute, 74.5/39.1\% Vs.\ Empty,
and 92.5/83.1\% pair separation. The Single--Dual gap reflects the joint
criterion and possibly different state/mode difficulty. Thus the complete
coordinate pathway improves prompt sensitivity and directional correctness,
including compatible two-component requests. The following representation analysis checks that its
intended code geometry is retained rather than assigning the effect to one
module.

\subsection{Latent-Space Analysis}

Figure~\ref{fig:codebook-tsne} visualizes 900 single-atom and 900
compatible-dual $d_M^a$ directions per arm. Before t-SNE, in the original
512-D space, the labeled atom is nearest for every single sample, and the target
pair is the two nearest codes for 99.1\%/99.7\% of right/left dual samples.
Mean angular errors
are 4.96$^\circ$/6.37$^\circ$ for single atoms and 14.91$^\circ$/11.84$^\circ$
to the normalized weighted Top-2 target for duals. Mean codebook separation is
94.61$^\circ$/94.66$^\circ$, indicating no collapse.
The t-SNE panels are qualitative; nearest-code and angle measures are computed
before projection.

With observation, VLM prefix cache, diffusion state, and noisy action fixed,
replacing spherical $z_M$ by its nominal direction changes the final
action-expert hidden state by 8.9/10.1/10.3\% RMS on Fruit/Plug/Vase;
batch-shuffling $z_M$ changes it by 12.7/12.8/14.0\% (32 native-subtask samples
per domain). Thus the spherical coordinate materially modulates this representation.

\subsection{Closed-Loop Task Progress}
\begin{figure}[H]
\expandafter\def\csname @captype\endcsname{figure}
\centering
\begin{minipage}[t]{0.48\linewidth}
\centering
\includegraphics[width=\linewidth]{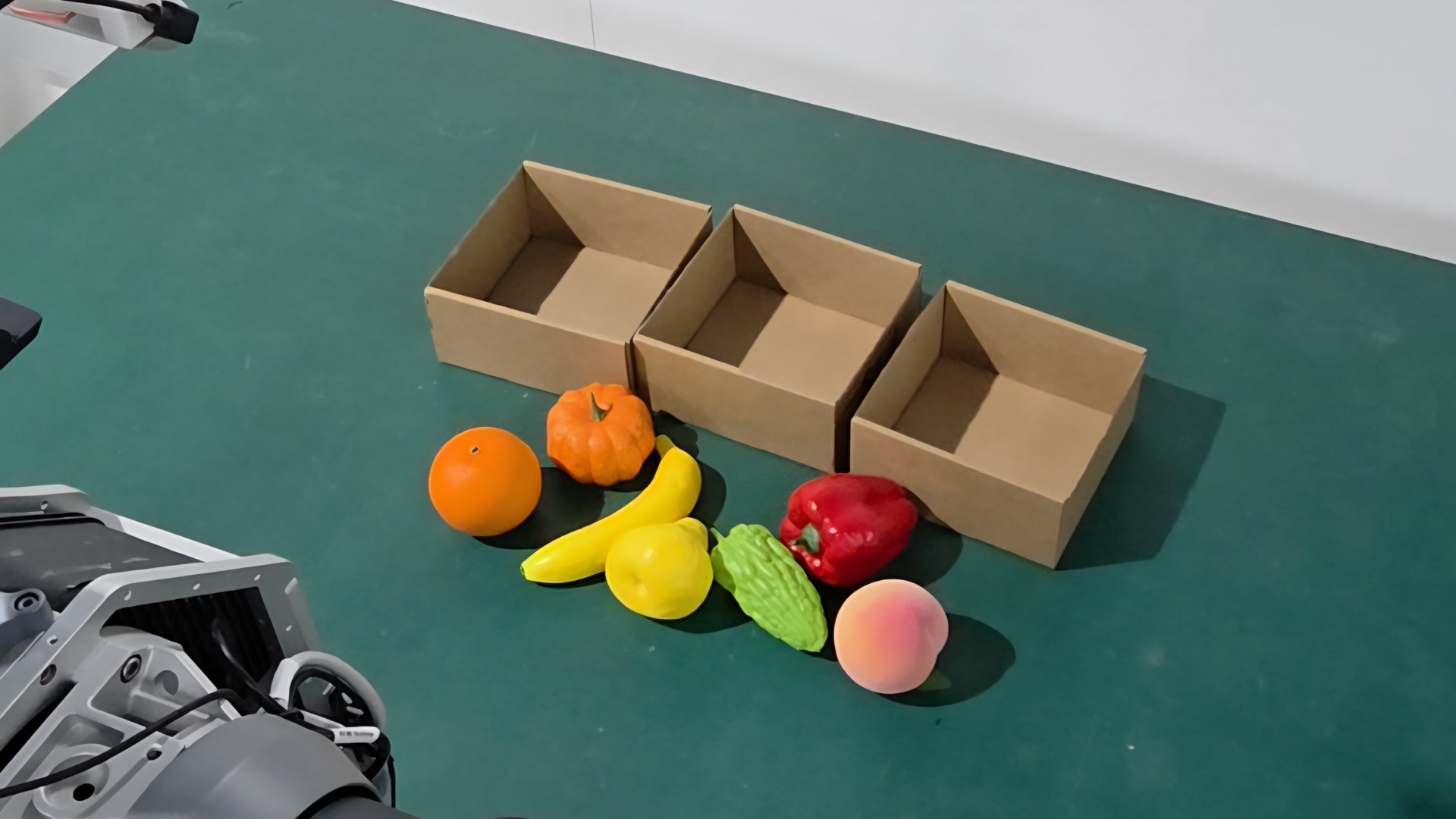}\\[-0.5mm]
{\scriptsize (a) Multi-fruit sorting}
\end{minipage}\hfill
\begin{minipage}[t]{0.48\linewidth}
\centering
\includegraphics[width=\linewidth]{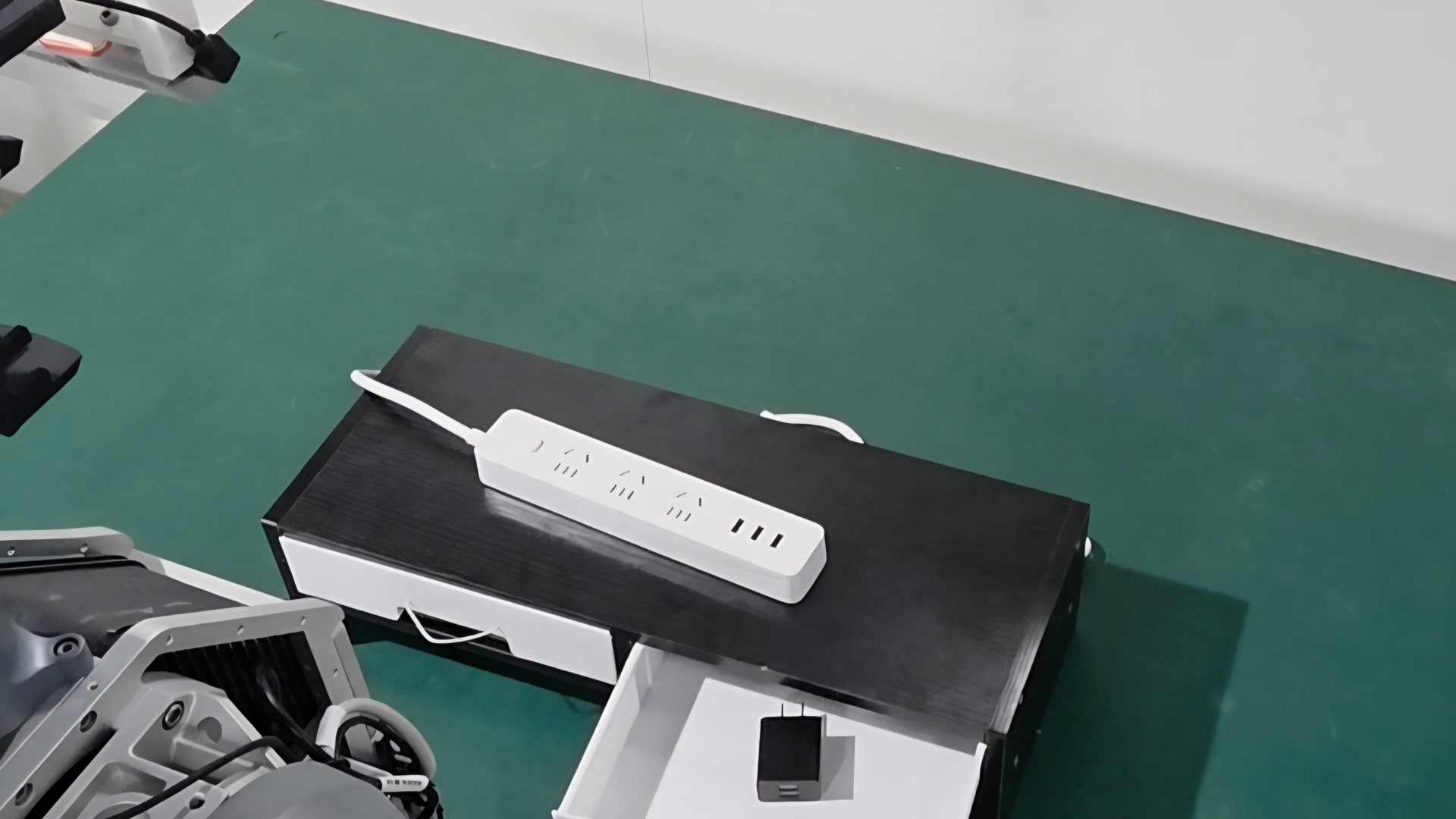}\\[-0.5mm]
{\scriptsize (b) Drawer retrieval and plug insertion}
\end{minipage}\\[7pt]
\begin{minipage}[t]{0.48\linewidth}
\centering
\includegraphics[width=\linewidth]{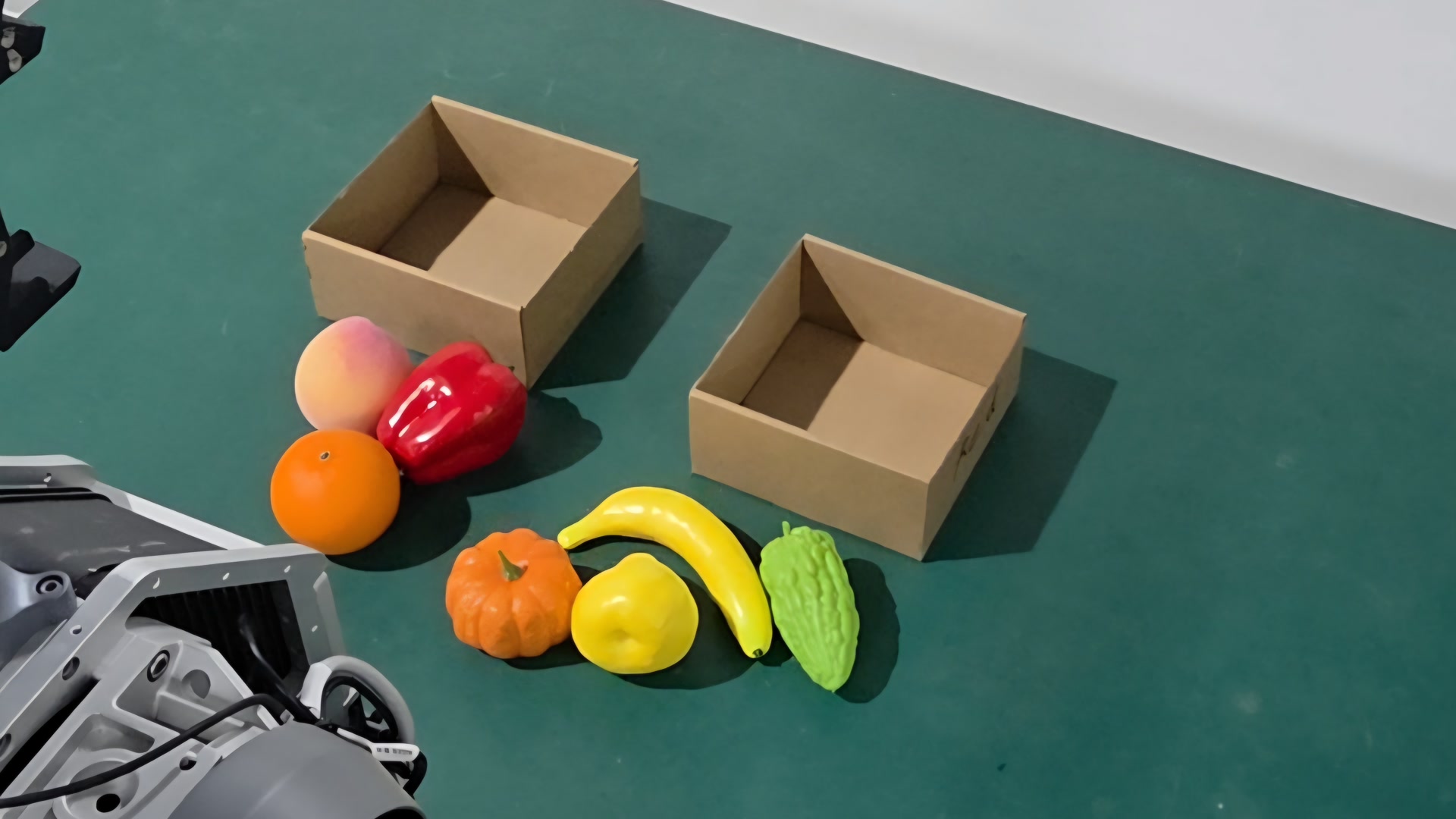}\\[-0.5mm]
{\scriptsize (c) OOD fruit ordering}
\end{minipage}\hfill
\begin{minipage}[t]{0.48\linewidth}
\centering
\includegraphics[width=\linewidth]{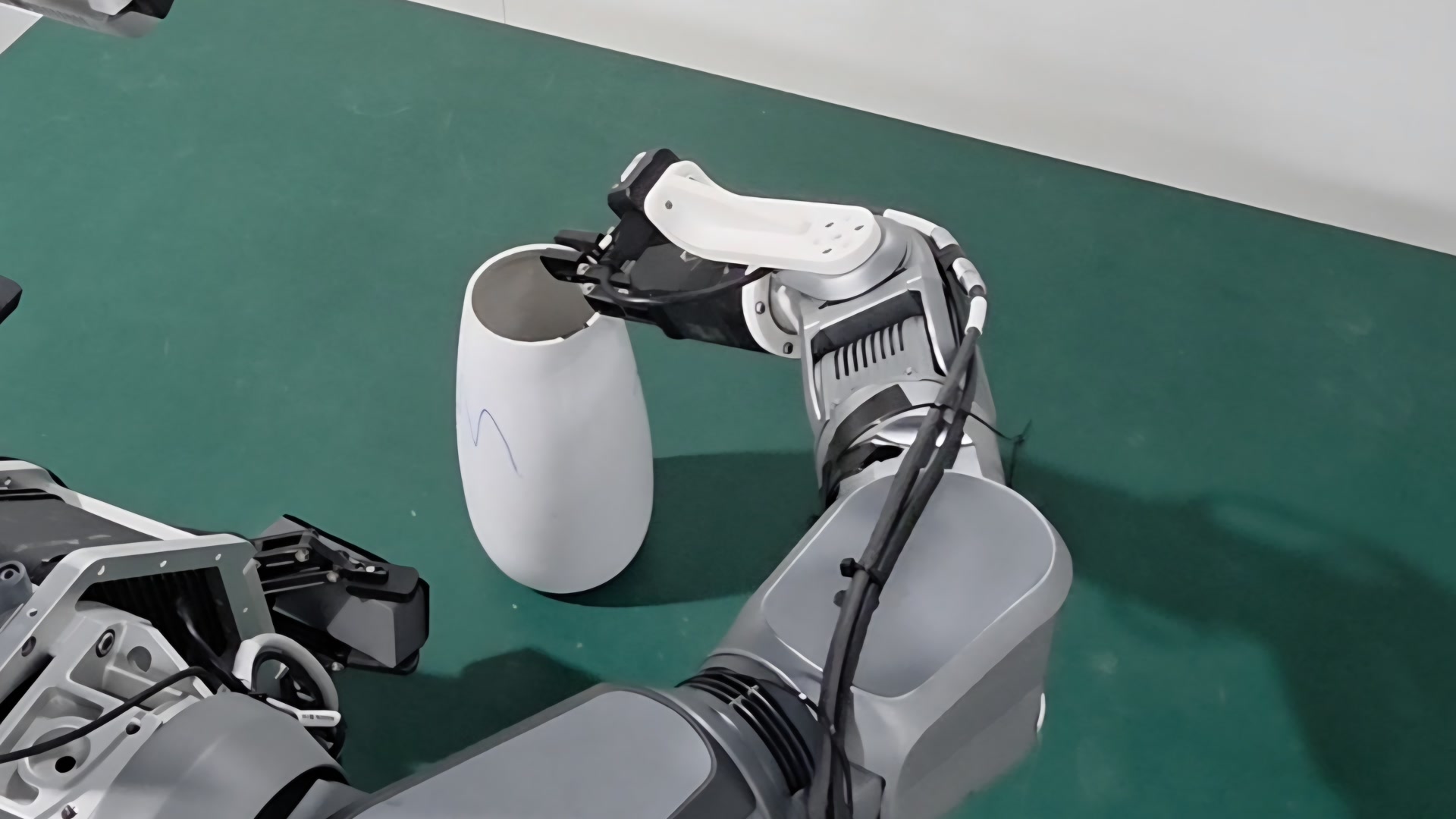}\\[-0.5mm]
{\scriptsize (d) Bimanual vase wiping}
\end{minipage}
\caption{Real-robot setups for the four experiment families. The two fruit
scenes differ: (a) measures closed-loop subtask steering under human-provided prompts with randomized familiar positions;
(c) uses the same protocol with unseen fruit placement and target-box positions.}
\label{fig:tasks}
\end{figure}

Figure~\ref{fig:tasks} shows the four closed-loop scenes and
Table~\ref{tab:main} reports their task progress. The fruit families test whether
object- and destination-specific subtasks can use per-arm motion trends under
visual context. Unlike Table~\ref{tab:steering}, these trials prescribe
neither an atomic axis nor an endpoint: the policy must identify the named
object or destination from images and compose the corresponding trends across
successive horizons. Thus fixed-state geometric redirection in
Table~\ref{tab:steering} is tested here as object- and destination-level
subtask composition into scene-conditioned task progress.
Fine-tuned $\pi_{0.5}$ retains a competent visual motor
prior (82.7/60.5\% familiar/OOD), but its lower prompt-contrast scores indicate
limited redirection of that prior when target positions and boxes change.
\method{} preserves strong visual performance while exposing an FK-grounded
per-arm coordinate; progress rises to 88.9/87.8\%.

\begin{table}[H]
\footnotesize
\centering
\caption{Mean normalized task progress (\%) over 50 trials on the four
real-robot setups in Fig.~\ref{fig:tasks}; ``--'' denotes untested pairs.}
\label{tab:main}
\setlength{\tabcolsep}{3pt}
\begin{tabular}{lcccc}
\toprule
Method & Fruit & Fruit (OOD) & Plug & Vase \\
\midrule
Fine-tuned $\pi_{0.5}$ & 82.7 & 60.5 & 55.0 & 65.2 \\
RSS & 54.5 & 39.1 & -- & -- \\
LA4VLA-style & 60.7 & 40.4 & -- & -- \\
ForceVLA & -- & -- & 30.0 & 40.4 \\
\method{} & \textbf{88.9} & \textbf{87.8} & 59.0 & 71.6 \\
\method{} + Force & -- & -- & \textbf{78.5} & \textbf{75.2} \\
\bottomrule
\end{tabular}
\end{table}

The two fruit baselines fail differently. LA4VLA-style improves offline prompt
contrast, but its language--direction binding is not tied to a fixed geometric
coordinate after full-observation training; its limited Absolute gains and
60.7/40.4\% execution are consistent with visual-control interference. RSS has
the strongest baseline single-atom separation, but one action-distribution
residual rather than independent bilateral coordinates; its weak dual steering
and 54.5/39.1\% fruit progress illustrate that separable prompt responses alone do
not guarantee executable subtask composition.

Relative to fine-tuned $\pi_{0.5}$ (55.0/65.2\%), force-free \method{} already
reaches 59.0/71.6\% on Plug/Vase, while \method{} + Force with matched history
raises these to 78.5/75.2\%. Plug gains more (19.5 points), consistent with discrete drawer,
alignment, and insertion contacts benefiting from RTC suffix correction. On
Vase, force-free \method{} already reaches 71.6\%, indicating that visual context
and the per-arm action composition supplied by \method{} account for much of
task execution; matched force history then yields a smaller, but positive,
3.6-point incremental gain. ForceVLA reaches 30.0/40.4\% in our matched reproduction; this result is
consistent with, but does not establish, a mismatch between direct force routing
and the pretrained visuomotor prior. \method{} instead makes a bounded,
non-accumulating update to its motion coordinate and regenerates only the suffix;
Table~\ref{tab:force-ablation} distinguishes current contact from phase.

With non-force inputs fixed, shifted history breaks current-contact alignment
but retains RTC phase, identifying the execution stage for segment correction.
This helps explain its gain over the disabled route. Matched history further
reduces joint/TCP RMSE by 24.0\%/21.3\% relative to shifted history,
highlighting the value of aligned force feedback.

Planner reliability is evaluated separately at subtask boundaries: without
task-specific SFT, Qwen3.5-4B produces the correct next subtask within five
successive 1~Hz queries in 94.6\%/92.4\% of 500 Vase/Plug cases. This
boundary-level score does not measure end-to-end task success.

\begin{table}[H]
\centering

\caption{Force ablation on 50 Plug/Vase trials (suffix-weighted RMSE; lower is better).}
\label{tab:force-ablation}
\setlength{\tabcolsep}{3.6pt}
\footnotesize
\begin{tabular}{lcc}
\toprule
Force condition & Joint (rad) & TCP (mm) \\
\midrule
Route disabled       & 0.0581 & 21.01 \\
Half-episode shifted history & 0.0459 & 19.41 \\
Matched force history & \textbf{0.0349} & \textbf{15.27} \\
\bottomrule
\end{tabular}
\end{table}

\section{LIMITATIONS AND CONCLUSION}
\label{sec:conclusion}
The thirteen atoms specify dominant base-frame trends, not complete trajectories:
grasp configuration and fine motion remain with the action expert. Generalization
depends on semantic and action support in pretraining and demonstrations, and
FK-derived labels inherit calibration error. Force adapts generated motion but
does not explicitly regulate wrench or guarantee contact stability; planner
reliability is boundary-level rather than end to end.

Within these limits, \method{} organizes existing semantic and action support
into named per-arm motion coordinates that condition every action-expert block.
A bounded, non-accumulating force update revises these coordinates during
contact while retaining the nominal semantic anchor. Our evaluations support
improved single- and compatible dual-direction steering, closed-loop fruit
subtask steering, and contact-rich task progress, providing an interpretable
interface for language guidance and force-conditioned motion adaptation.

\bibliographystyle{IEEEtran}
\bibliography{references}

\end{document}